\documentclass[letterpaper,twocolumn,fleqn]{article} 

\usepackage{ist}
\usepackage{times}
\usepackage{epsfig}
\usepackage{graphicx}
\usepackage{amsmath}
\usepackage{amssymb}
\usepackage{tabularx}
\usepackage{multirow}

\title{The Blessing and the Curse of the Noise behind Facial Landmark Annotations$^*$}

\author{Xiaoyu Xiang$^a$, Yang Cheng$^a$, Shaoyuan Xu$^a$, Qian Lin$^b$, Jan Allebach$^a$ \newline
$^a$School of Electrical and Computer Engineering, Purdue University, West Lafayette, IN 47907, USA \newline
$^b$HP Labs. Palo Alto, CA 94304, USA\newline}

\date{} 

\begin{document} 

\maketitle 

\thispagestyle{empty} 


\begin{abstract}
\let\thefootnote\relax\footnote{$^*$Research supported by HP Inc., Palo Alto, CA.}The evolving algorithms for 2D facial landmark detection empower people to recognize faces, analyze facial expressions, etc. However, existing methods still encounter problems of unstable facial landmarks when applied to videos. Because previous research shows that the instability of facial landmarks is caused by the inconsistency of labeling quality among the public datasets, we want to have a better understanding of the influence of annotation noise in them. In this paper, we make the following contributions: 1) we  propose two metrics that quantitatively measure the stability of detected facial landmarks, 2) we model the annotation noise in an existing public dataset, 3) we investigate the influence of different types of noise in training face alignment neural networks, and propose corresponding solutions. Our results demonstrate improvements in both accuracy and stability of detected facial landmarks.
\end{abstract}

\section{Introduction}

2D facial landmark detection is a fundamental technology behind face recognition \cite{liu2019faceset}, expression recognition \cite{xu2019real-time}, augmented reality 3D mask rendering, etc. Besides precise and accurate localization, there are several problems that have attracted more and more attention in recent years: landmark stability, facial landmark detection under extreme conditions such as occlusion, rare angle faces, low luminance, large movement, etc. Datasets play a pivotal role in addressing these problems in landmarks localization, including the early AFLW dataset \cite{koestinger2011annotated} with 21 point markup and the more recent 300-W \cite{sagonas2013300},  300-VW \cite{chrysos2015offline}\cite{shen2014unsupervised}\cite{trigeorgis2016mnemonic} with 68 point annotations, and WFLW \cite{wayne2018lab} with 98 manually annotated landmarks. These datasets have not only grown larger in scale, but also have become more diverse in attributes, including occlusion, pose, make-up, illumination, motion, and facial expressions. 

Although the scale of facial landmark datasets is growing, it is still far from being comparable with the tremendous size of face recognition datasets such as Ms-Celeb-1M \cite{guo2016ms}, which consists of 10M images of 100K celebrities. These large scale datasets, along with research on data cleaning \cite{wang2018devil}, drive the development of new methods to achieve better results in face recognition. The main factor that constrains the scale of facial landmark datasets is that, in the current stage, the labeling of landmarks heavily relies on manual annotation and verification. Different from establishing face recognition datasets whose labels can be cleaned automatically, building a facial landmark dataset is tedious and time-consuming. Besides, \cite{dong2018supervision} noticed that human annotations inherently have flaws in precision and consistency: the positions of the same landmark point annotated by different people vary a lot, even those of the points with clear features (e.g., corner of mouth). The variance in the training data would degrade the stability of landmark detectors, thus leading to perceptually unpleasant jitters when the detector is applied to videos. Traditional landmark detection frameworks treat each frame of a video as an individual input and pay little attention to temporal consistency; as a result, it is hard for the output points to maintain a consistent visual presence in consecutive frames.

Considering the fact that the inconsistency of landmark annotations widely exists in popular facial landmarks datasets, such as 300-W and 300-VW, we regard the unwanted jitters as a type of noise and design relevant experiments around this concept. 

The first goal of our work is to understand the noise in facial landmarks and how it will influence the training of deep convolutional neural networks (CNN). In other words, we are concerned with the relationship between the training set's noise and the model's performance, the extent that the noise can be reduced, and the best method to get a clean output. To achieve this goal, we propose two plausible metrics that measure landmarks' stability. Although we have to admit that currently it is almost impossible to get noise-free landmark points in either the training or the detection stage, a better understanding of the concerns expressed above would be beneficial for the design of more robust algorithms for real-world applications. 

The second goal of this paper is to propose a complete work-flow to help decrease the inconsistency of facial landmark annotations, that exists in a wide range of popular public datasets. In this paper, we show that training on a corrected dataset can actually improve the detector's performance. Our proposed work-flow can also boost the performance of existing methods. The corrected dataset used in training our facial landmark detector consists of approximately 4,000 still images, and 300 videos of 200 identities. Due to the nature of the dataset source, these images exhibit great variations in scale, pose, lighting, and occlusion. For a better comparison, we also carried out experiments on corrupted datasets by injecting noise on landmark annotations following previous research \cite{dong2018supervision}. By controlling the amount of additive noise, this study helps us to understand quantitatively the noise's influence on the detection accuracy. 


\section{Related Work}
\textbf{Semi-automatic Facial Landmarks Annotation:} To aid the manual annotation work, Christos et al. \cite{sagonas2013semi} proposed a semi-automatic methodology for facial landmark annotation in creating massive datasets. They used the annotated subset to train an Active Orientation Model (AOM) that provides an initialization to non-annotated subsets, and then classifies the results to ``good" and ``bad" manually. However, this kind of method can only reach a relatively accurate annotation by cleaning out the obviously ``bad" examples, but cannot avoid the jitters among different annotations.

\textbf{Face Alignment and Tracking:} To improve face alignment in video, Peng et al. \cite{peng2015piefa} proposed an incremental learning method for sequential face alignment. To better make use of the temporal conherency in image sequences, Peng et al. \cite{peng2016recurrent} designed a recurrent encoder-decoder network
model for video-based face alignment, where the encoding module projects the input image into a low-dimensional feature space, and the decoding module maps the features to 2D facial point maps. The recurrent module demonstrates improvements to the mean and standard deviation of errors by taking consideration of previous observations. However, this module unavoidably increases the network complexity and training difficulty. Besides, many researches applied tracking as an extension of face alignment, though it always results in drifting and loss of accuracy of facial landmarks. Conducting face tracking usually involves generic facial landmark detection, the combination of model-free tracking and re-initialization \cite{chrysos2018comprehensive}. Khan et al \cite{khan2017synergy} proposed a synergistic approach to eliminate tracking drifts, in other words, to apply face alignment when drifting happens. Still, stepping happens when shifting between tracking and detection results. Barros et al. \cite{barros2018fusion} applied a Kalman filter to fuse the tracking and detection.

\section{Methodology}
\subsection{Noise Modelling}

\subsubsection{Noise in Public Datasets}
Naturally, people assume that the ground truth in popular public datasets is accurate and precise. However, Dong et al. \cite{dong2018supervision} has noticed the existence of hand annotation noise. Here, we will look into the two most commonly used datasets\textemdash 300-W \cite{sagonas2013300} and 300-VW \cite{chrysos2015offline}\cite{shen2014unsupervised}\cite{trigeorgis2016mnemonic}:

\textbf{300-W} provides 68 2D landmark annotations for 3,837 face images. These images were split into four sets respectively: training, common testing, challenging testing, and full testing.  In this paper, our base detector was trained on the 300-W training set. In addition, we evaluated the detector's performance and modeled the output noise on the 300-W test set. 

\textbf{300-VW} is a video dataset consisting of 50 training videos with 95,192 frames. Its test set contains three subsets (1, 2, and 3) with 62,135, 32,805, and 26,338 frames, respectively. Among the three categories, Subset-3 is the most challenging. We apply the proposed method to correct landmarks of the 300-VW full set and report results on all three subsets.

\begin{figure}[htbp]
\begin{center}
\includegraphics[width=\linewidth]{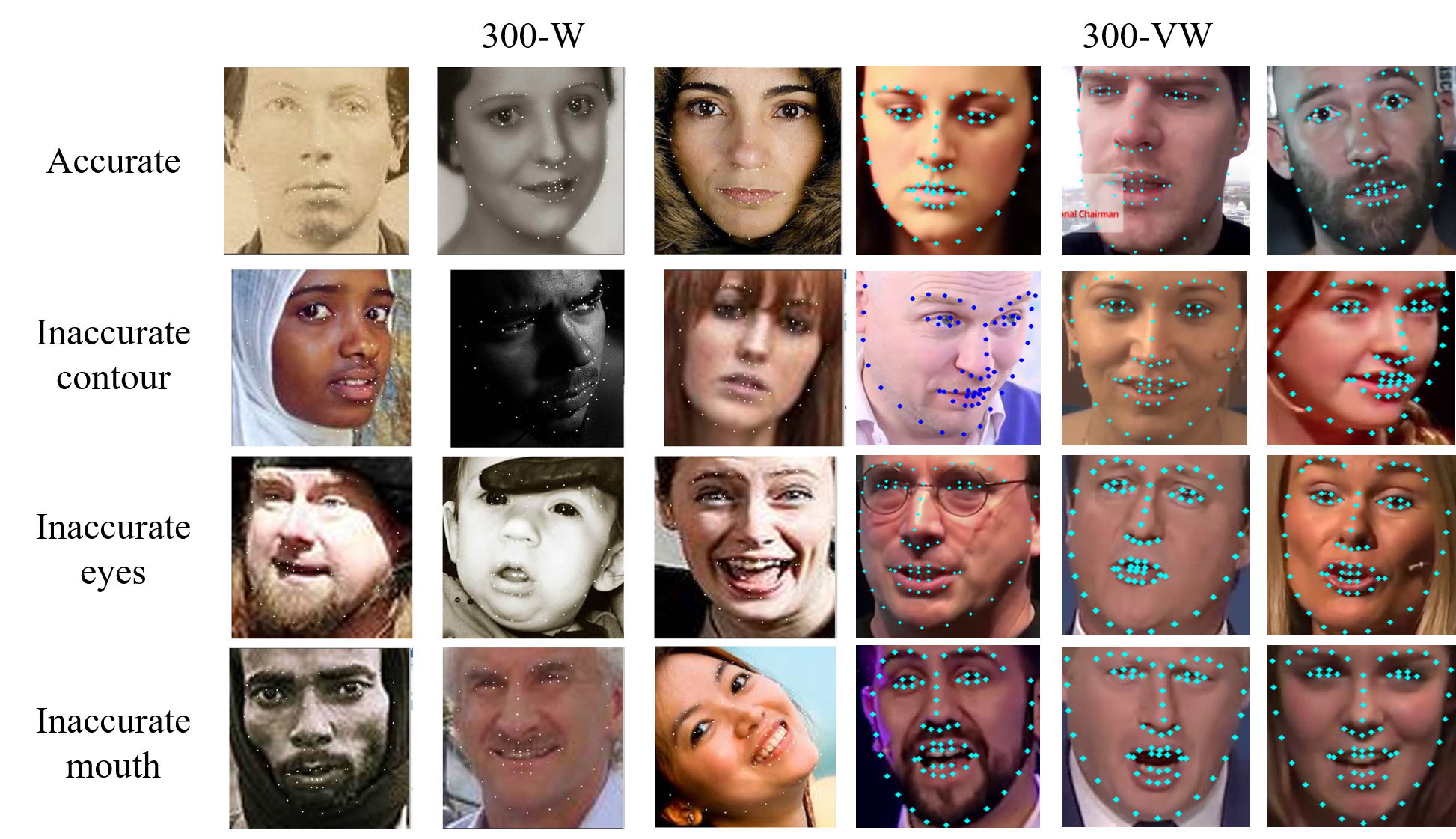}
\end{center}
   \caption{Noisy annotations in public datasets. \textmd{The images in the left 3 columns are from 300-W, and the images in the right 3 columns are from 300-VW. The quality of annotations is not consistent among these two well-known datasets. The reader is advised to zoom in to see the annotations.}}
\label{fig:inaccurates}
\end{figure}

Figure \ref{fig:inaccurates} shows some examples of inaccurate annotations in the 300-W and 300-VW datasets. Images in the dataset can be categorized into four categories: \textbf{accurate subset}: data samples with acceptable ground truth; \textbf{inaccurate eyes}, \textbf{inaccurate mouth}, and \textbf{inaccurate contour}. Besides, there are also images with more than one inaccurate facial component. 

\subsubsection{Metrics}
\label{noise_metric}
In this section, we will introduce the widely used metric for facial landmark accuracy, and our proposed methods to measure the stability of detected facial landmarks quantitatively.

\textbf{Accuracy} reflects the difference between the predicted result and the ground truth. A good facial landmark detector should produce results of low prediction error for any given inputs, including still images and video frames.

The normalized mean error (NME) is widely used as the evaluation metric for accuracy. It is defined as follows:
\begin{equation}\label{eq:nme}
    \text{NME} = \dfrac{\frac{1}{N}\sum_{i=1}^{N}L_2(f^{\theta}_i(x), \hat{y}_i)}{d},
\end{equation}
where $\hat{y}_i$ and $f^{\theta}_i(x)$ denote each ground truth point based on human annotation and the predicted point, respectively; and $x$ is the input image. $N$ is the number of landmarks on a face (in this paper, $N=68$), and $d$ represents the outer ocular distance (the distance between outer corners of eyes) for normalization. It is worth noting that in some previous papers, the distance between centers of eyes (inter pupil distance) was also used for normalization. 

\textbf{Precision} reflects the robustness of a model when given inputs with different kinds of noise, e.g. pixel-wise noise including camera shot noise, Gaussian blur, or re-centering, etc, which can exist in video frames. These kinds of noise will not change the spatial distance among different facial components of ground truth, but may cause jitters in detection outputs. Since the absolute locations of facial landmarks should be unchanged, we can define the variance of detected points for precision as follows:

\begin{itemize}
    \item Standard Deviation (STD). This metric doesn't require annotated ground truth but a test set of still images, e.g., video frames of an unmoved face from a fixed camera. In this case, each frame is naturally injected with different camera shot noise. The normalized standard deviation of the landmark locations in the same video can be used as a metric for stability:

\begin{equation}
    STD_i = \dfrac{1}{d}\sqrt{\dfrac{\sum_{j=1}^{n}(f^{\theta}_i(x_j) - \overline{f^{\theta}_i(x)})^2}{n-1}},
\end{equation}
where $j$ refers the index of frames: $1,...,j,...n$; and $\overline{f^{\theta}_i(x)})^2$ is the position of the $i$-th landmark averaged over the $n$ frames. The output of this equation is the standard deviation of $i$-th landmark coordinate. Larger standard deviation indicates more jitters.

\item Standard Deviation of Difference (SDD). This metric requires annotated ground truth as a reference: Firstly, the difference between the ground truth and detection results for each frame is calculated as $\Delta y_i = f^{\theta}_i(x) - y_i $. In the ideal case where the detection results exactly follow ground truth, the variance of difference should be low no matter how big the difference is. Larger SDD values indicate more jitters. For a given video of frames $j\in \{1,\dots,n\}$, the formula for this metric is given by:

\begin{equation}\label{eq:sdd}
    SDD = STD(\Delta y) = \dfrac{1}{d} \sqrt{\dfrac{\sum_{j=1}^{n}(\Delta y_j - \overline{\Delta y})^2}{n-1}}
\end{equation}


\end{itemize}

\begin{figure*}[htbp]
\begin{center}
\includegraphics[width=\linewidth]{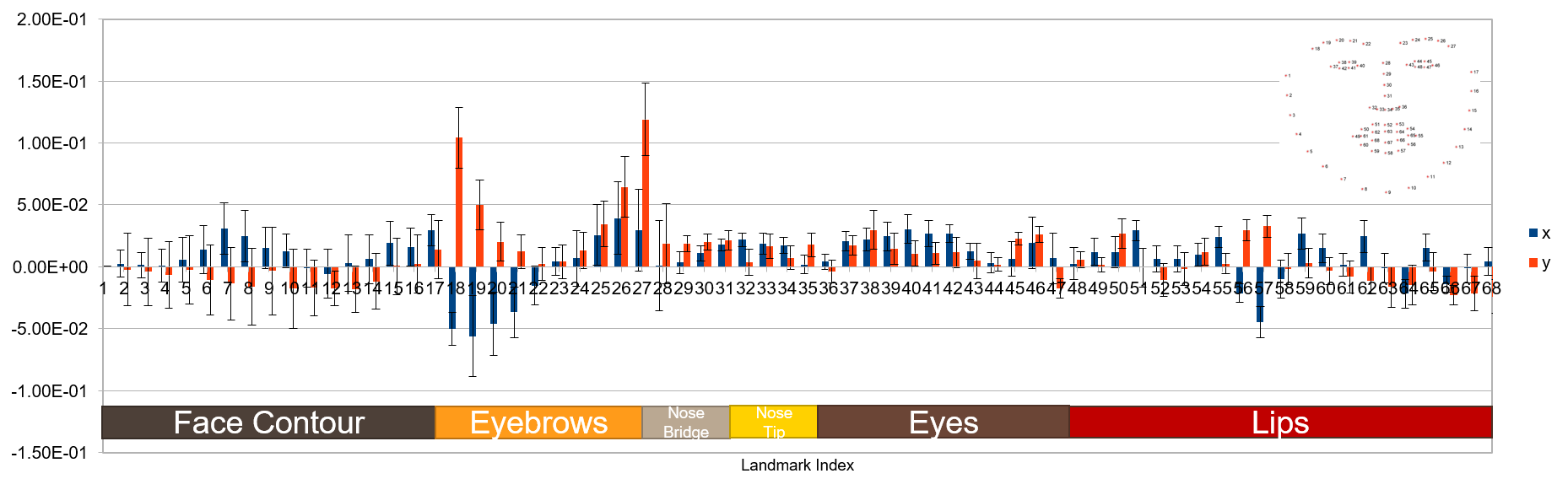}
\end{center}
   \caption{Example of the detection noise (SDD (Equation \ref{eq:sdd})) in $X$ (blue bar) and $Y$ (red bar) coordinates. The horizontal axis denotes the index number of the facial landmark, e.g. points 1 $\sim$ 17 represent face contour points, 18$\sim$27 stand for eyebrow points, etc. The solid bars are the mean values of the detection results' difference, and the whiskers represent the standard deviation of the difference as defined before. Bigger error bars of points indicate that these points are unstable in that direction, while mean values can tell us the prediction error, or ``bias" from ground truth. Ideally, we wish the model's prediction result to be an unbiased estimate of the ground truth, which corresponds to zero mean values in this graph.}
\label{fig:noise_std}
\end{figure*}

In a similar manner, we can then define the detection noise as the deviation between the detection results and the target values, as mentioned before. We can use the distribution of the detection noise as a tool to visualize the detector's robustness and stability. Figure \ref{fig:noise_std} shows an example of the noise of each landmark point in both the $X$ and $Y$ coordinates that are calculated from a ``pseudo" video generated by the augmentation method mentioned later in Section \textit{Experimental Setup}. In this figure, the detection noise is plotted in bar graphs with error bars, where blue and red bars correspond to the mean value of the detection results' difference from ground truth in the $X$ and $Y$ directions, respectively. The error bars are calculated from the standard deviation of these differences among a group of test images, as defined previously in this paper. We can get some interesting information from this plot: the most unstable points are the eyebrows' ends, and then the lowest point of the lips, which are actually the most deformable points on our face. Besides, the color fading of the eyebrows' ends also throws challenges to the landmark detector.

\begin{figure*}[htbp]
\begin{center}
\includegraphics[width=\linewidth]{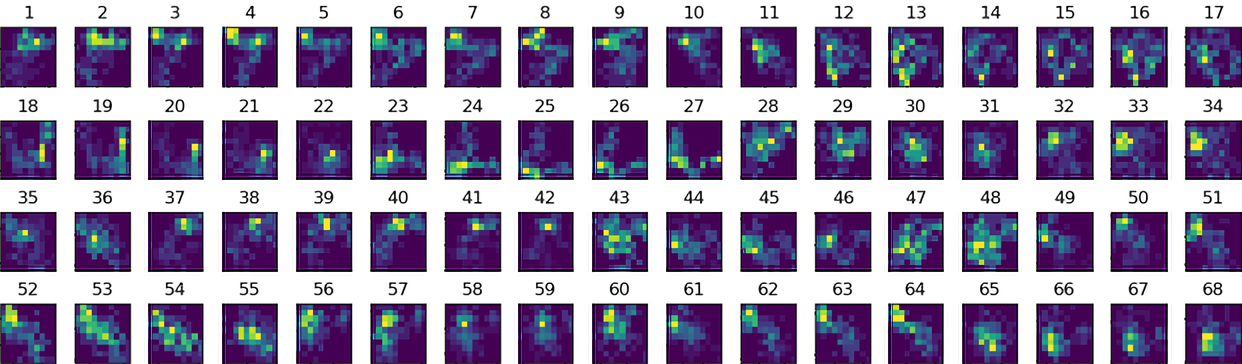}
\end{center}
   \caption{Example of the noise (SDD (Equation \ref{eq:sdd})) plotted as a 2d histogram. Each histogram represents the noise distribution of every facial landmark point in the $X$ and $Y$ coordinates. Ideally, a stable point should have a Gaussian distribution with a clear peak. If the prediction result is an unbiased estimate of the ground truth, this peak should be located at the zero point of the histogram in both $X$ and $Y$.}
\label{fig:noise_hist}
\end{figure*}

Figure \ref{fig:noise_hist} shows the detection noise plotted as a 2d histogram with both $X$ and $Y$ directions, which serves as another way to visualize the spatial distribution of the noise. In previous research papers, all noises are assumed to be of the Gaussian distribution for simplicity. In this graph, the scattered map of points without clear peaks usually indicates that the noise severely deviates from a Gaussian distribution. This graph can be used to select stable landmarks as reference points for further face alignment in Augmented Reality (AR) applications.

The noise in the detection result is connected with inaccurate locations and jarring visual effects. Thus, methods including tracking and temporal filter are applied in post-processing to reduce the noise as discussed in Section \textit{{Methods to Reduce Noise}}. Although the physical noise cannot be eliminated due to the limitation of the input data, it is possible to attain better results using the prior knowledge about the noise distribution. 
\subsubsection{Theoretical Assumptions}
\label{noise_theory}

Assume we have a set of unreliable annotations $(\hat{y_1}, \hat{y_2}, ...)$ of facial landmarks. Denoting the ``true and only" landmark coordinate as $y$, we describe their relationship by the following equation:

\begin{equation}
\hat{y} = y + \delta(y),    
\end{equation}
where $\delta(x)$ is the difference between the annotated data and the true value, or to say, the human annotation noise. Usually, the noise distribution is assumed to have zero mean and finite variance \cite{mccullagh1989generalized}:

\begin{equation}
    E[\delta(y)] = \sum_{i=1}^{n} \delta(y_i)p_i = 0, n\rightarrow \infty
\end{equation}

Based on this assumption, we can get

\begin{equation}\label{eq:exp_yi}
    E[\hat{y}] = E[y + \delta(y)] = y + E[\delta(y)] = y,
\end{equation}
indicating that the expected value of the human annotated value is the true unknown $y$. In practice, it is possible to approach $y$ through a learned mapping function with a sufficient number of randomly distributed training samples. Most facial landmark models are trained with L2 loss (MSE Loss): 

\begin{equation}
    L2 Loss = \dfrac{1}{2N}\sum_{i=1}^{N}||f^{\theta}_i(x) - \hat{y}_i||^2_2 ,
\end{equation}
where $x$ denotes the input image data, and $f^{\theta}$ is the mapping function. Thus, $f^{\theta}_i(x)$ is the network's output for the $i$-th landmark, $\hat{y}_i$ is the value of the ground truth for $i$-th landmark based on human annotation, and $N$ refers to the number of elements in the network's output array. For the common case of 68-point facial landmarks, $N=136$.

The optimization goal of L2 Loss is given by:

\begin{equation}
\begin{array}{l}
    argmin_{\theta} E_{\hat{y}_i}[(f^{\theta}_i(x) - \hat{y}_i)^2] = \\ 
    argmin_{\theta} E_{f^{\theta}_i(x)}\left[[E_{\hat{y}_i | f^{\theta}_i(x)} [(f^{\theta}_i(x) - \hat{y}_i)^2]\right]
    \end{array}
\end{equation}

As mentioned before, the noise in landmark positions is assumed to be zero-mean with finite variance for all images in our dataset. A network trained with L2 Loss, from a statistical point of view, will produce outputs that approximate the conditional mean of the target values in the training set:

\begin{equation}
    E[\hat{y}_i| f^{\theta}_i(x)] = E[\hat{y_i}]
\end{equation}

As defined in Equation \ref{eq:exp_yi}, the expectation of $\hat{y}_i$ can approach the unobserved true value $y_i$ if there are enough random inputs. 

However, if $\delta(y)$ belongs to other types of noise, L2 loss may not guarantee that the destination of optimization is $E[\hat{y}_i]=y_i$. For example, if injected with salt-and-pepper noise (impulse noise), the conditional median would be better at approaching the true value, which can be learned by least-absolute-value training with L1 Loss. Following a derivation similar to that above, we can get the correspondence between the loss function and the noise type shown in Table \ref{tb:noiseloss}.

\begin{table}[htbp]
\caption{Table 1. Loss function for each type of noise}
\label{tb:noiseloss}
\begin{center}
\begin{tabular}{l c}
\hline
Noise Type & Loss Function \\
\hline
Additive Gaussian noise  &  \multirow{2}{*}{L2 loss} \\
Poisson noise &  \\
Bernoulli noise(binominal noise) & Modified L2 loss \cite{ulyanov2018deep} \\
Salt-and-pepper noise & L1 loss\\
Random-valued impulse noise & L0 loss \\
\hline
\end{tabular}
\end{center}
\end{table}

Researchers from the image denoising area have already noticed the connection between noise type and training loss type \cite{ulyanov2018deep}. In \cite{lehtinen2018noise2noise}, the model learns to restore images by only looking at corrupted data with synthetic noise. Previous research \cite{feng2018wing} also compared training facial landmark detectors using L1 and L2 loss, and a custom loss function. The different performances of the acquired models, along with our noise modeling results indicate that the actual noise in public training data is more complex than that based on theoretical assumptions. 

\subsection{Methods to Reduce Noise}
\label{noise_reduce_methods}

Based on previous assumptions about noise, we propose a method to reduce the spatial noise with the temporal information from adjacent frames. For each image, we assume the relationship between the ground truth and the prediction is:

\begin{equation}
    gt_i = p_i + err_i ,
\end{equation}
where $gt$ denotes the ground truth, $p$ stands for the prediction(or annotation), and $err$ is the error for view $i$. The landmark points of different frames can be regarded as individual samples around the ground truth. The difficulty lies in calculating the errors from different frames under the same view. We choose optical flow to find the corresponding points between two images; this process can be expressed as follows:

\begin{equation}
    F_t(p_i) = F_t(p_i + err_i) = p_{i,k} ,
\end{equation}
where $F_t$ represents the point registration process, and $p_{i,k}$ stands for the information in sample $p_i$ under view $k$. Note that the acquired $p_{i,k}$ includes $err_{i,k}$ that is cast to the new view. In this way, we can apply the previous equation in Section \textit{Noise Modelling} to reduce noise.

\begin{figure*}[htbp]
\begin{center}
\includegraphics[width=\linewidth]{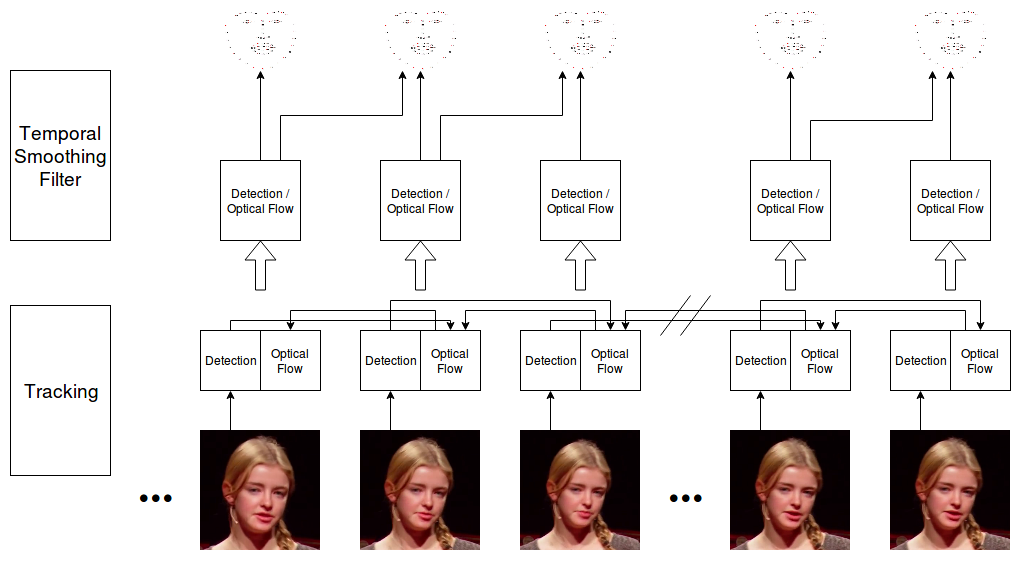}
\end{center}
   \caption{Overview of our framework. \textmd{Each frame of the video sequence is fed into the cascaded detection network to obtain the facial landmarks prediction. Then the optical flow algorithm is applied to the facial landmarks of each frame to predict the landmarks of neighboring frames. Each frame now has two sets of facial landmarks, which are then fused together by assigning different weights to the landmarks of these two sets. For each landmark, if the prediction from the optical flow is close to that from the detection network, a higher weight is assigned to the prediction from optical flow, and vice versa.}}
\label{fig:framework}
\end{figure*}

Until now, optical flow has been successfully used as a tracking method in many applications. Optical flow assumes that the brightness of a point that moves slightly from frame to frame does not vary as the time varies; and the movement of the neighbors follows in the same way. In our method (Figure \ref{fig:framework}), the 68 facial landmarks detected from the first frame are considered the initial points of interest to be tracked by the optical flow, which gives the predicted positions of these points in the second frame. Then, the optical flow is used again on these predicted points but in a reverse way, to predict their positions in the first frame. Finally, the detection model gives another set of 68 facial landmarks in the second frame. Therefore, in both frames, there exist two sets of 68 points, predicted by the detection model and the optical flow. To supervise the correctness of optical flow, the tracking result is chosen over the detection result in the second frame, if the tracking result is close to the detection result, and vice versa. 

Usually, the optical flow performs well in tracking, but there exist some cases that easily cause the failure of optical flow. For example, the point on the upper eyelid has been tracked stably by optical flow, but when the person blinks their eyes, the point may stick to the lower eyelid instead of going up with the upper eyelid. Therefore, we assume that the tracking result from optical flow should be close to the detection result. Also, the reason for applying optical flow again to the tracking points in the second frame to give their positions in the first frame is to check the trustworthiness of the optical flow. This is based on the assumption we made on the optical flow algorithm that the results of a two-frame scenario should be consistent, regardless of the sequence of these two frames. 

Although applying optical flow to the detection results improves the stability of the facial landmarks, it suffers from the failure case in which the detection result is not close to the tracking result; so it is hard decide which result should be chosen as the final landmark of the frame. To resolve this issue, we developed a simple but efficient way to smartly leverage the detection result and the tracking result. From a high-level view, the final facial landmarks are given by
\begin{equation}
    P_{final}=\alpha P_{detection} + \beta P_{tracking},
\end{equation}
where $\alpha$ and $\beta=1-\alpha$ denote the weights assigned to the detection result and tracking result, respectively. To determine the value of $\alpha$, we made two assumptions: 1) the landmarks in the first frame should be close to the landmarks in the second frame, predicted by applying optical flow to the landmarks in the first frame, and 2) the landmarks, obtained by applying optical flow to the landmarks in the first frame to the second frame and then back to the first frame, should be close to the original landmarks in the first frame. By measuring these two distances, the value of $\alpha$ can be calculated, since the larger the distances, the larger the value of $\alpha$, in other words, the less trustworthy the tracking result is.

\section{Experiment and Analysis}

\subsection{Experimental Setup}
\label{sec:exp_set}
\begin{figure}[htbp]
    \begin{center}
        \includegraphics[width=0.77\linewidth]{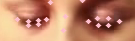}
    \end{center}
    \caption{Comparison between the results obtained from the single-level network and the cascaded network. \textmd{The landmarks on the left-side eye are predicted by the cascaded network, while the landmarks on the right-side eye are predicted by the single-level network. This indicates that the cascaded network performs better on the components, since it only focuses on them.}}
    \label{fig:ruiyi_out}
\end{figure}

\textbf{Baseline.}  In \cite{mao2018robust}, Mao et al. proposed a cascaded VGG-style network, which demonstrated a strong ability to detect facial landmarks accurately and performed extremely well on the 300-W test dataset \cite{sagonas2013300}. This cascaded network is designed as a two-level network, where the first level outputs initial 68 facial landmarks, and the second level further refines the prediction results for each component, e.g., eyes and mouth, by fusing the global information obtained by the first level network and the features extracted by the second level network. Compared to the conventional single-level network, this cascaded network outputs facial landmarks of higher accuracy. Figure \ref{fig:ruiyi_out} shows an example comparing the results obtained from the first-level network and the cascaded network \cite{mao2018robust}. In this paper, we adopt the first level of Mao's cascaded network as the base model.

We choose PyTorch \cite{NEURIPS2019_9015} to implement this work. We apply MSE loss to train the network and Stochastic Gradient Descent (SGD) as the optimizer. The initial learning rate is set to be 0.05, and the maximum number of epochs is 100 with a batch size of 512. Our network is trained on one Nvidia Quadro M6000. All input images are resized to $128 \times 128$ and normalized in RGB channels before sending them to the network.

\textbf{Fusion} By jointly detecting and tracking the landmarks using optical flow, we are able to reduce the instability of the landmarks that appeared in 300-VW. In each video, the ground truth landmarks can be treated as the detection result, then starting from the second frame, the tracking result is obtained by applying optical flow on the detected landmarks in the previous frame. Then these tracked landmarks in the second frame are used by the optical flow to predict the landmarks back to the first frame. As previously mentioned in Section \ref{noise_reduce_methods}, the value of $\alpha$ and the weight assigned to each detected landmark can be calculated to obtain the final facial landmarks in the current frame. 

\begin{figure}[htbp]
\begin{center}
    \includegraphics[width=0.385\linewidth]{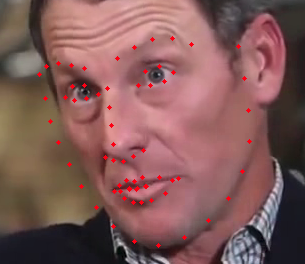}
    \includegraphics[width=0.38\linewidth]{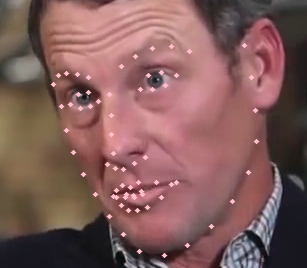}
\end{center}
    \caption{Comparison between the image from the original 300-VW dataset and our corrected 300-VW dataset. \textmd{The annotations of the original 300-VW dataset are not temporally consistent, causing the original dataset to be too noisy to be used. The image on the left side shows one of the examples: the landmarks on the contour are not closely attached to the face contour. However, this is not the case in the image on the right side, which is obtained from our corrected 300-VW dataset.}}
    \label{fig:good_bad}
\end{figure}

\textbf{Data Augmentation.} We design a data augmentation method especially for the video task. The goal of our augmentation is to turn a still image into a ``pseudo" video, with continuous changes in the pixel-wise noise, motion blur, brightness, scale, projective distortions, etc. The key point of doing this data augmentation is to acquire a training video without intra-frame noise in landmark locations. To generate diversified videos with these augmentation methods, we borrow the idea of a``storyboard" from designers to assign the start status and the end status of the generated video, where the intermediate steps are set to adapt to the fps (frame-per-second) in order to simulate real-world changes.

\begin{figure}[htbp]
\begin{center}$
\begin{array}{ccccc}
\includegraphics[width=0.15\linewidth]{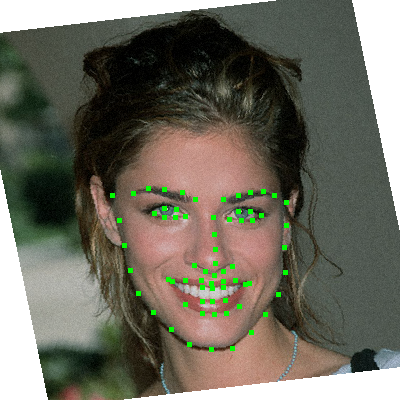}&
\includegraphics[width=0.15\linewidth]{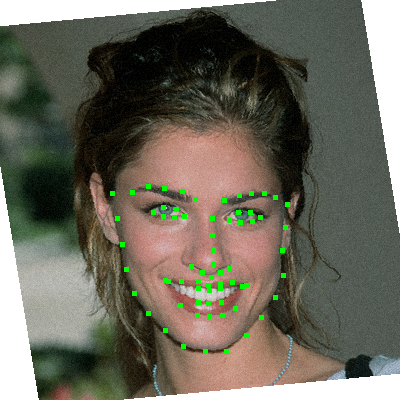}&
\includegraphics[width=0.15\linewidth]{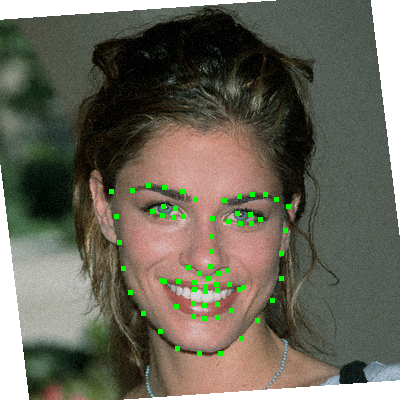}&
\includegraphics[width=0.15\linewidth]{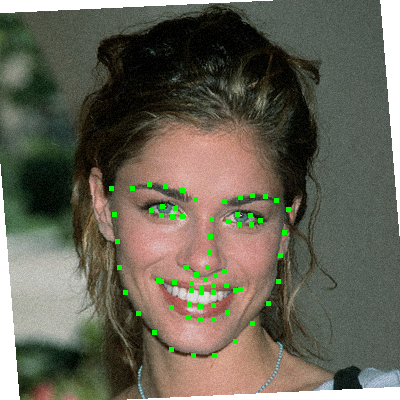}&
\includegraphics[width=0.15\linewidth]{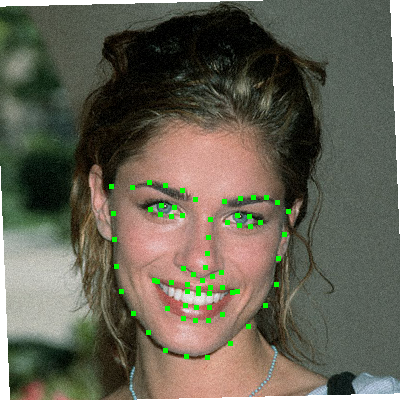} \\
\includegraphics[width=0.15\linewidth]{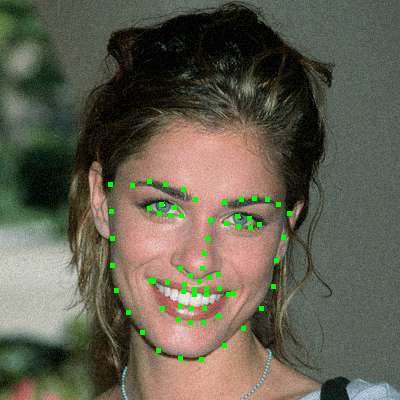}&
\includegraphics[width=0.15\linewidth]{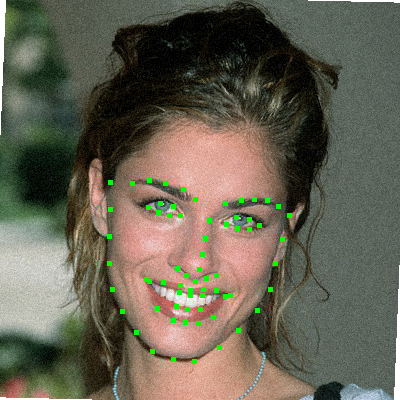}&
\includegraphics[width=0.15\linewidth]{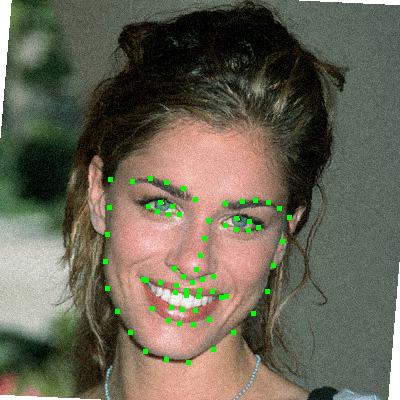}&
\includegraphics[width=0.15\linewidth]{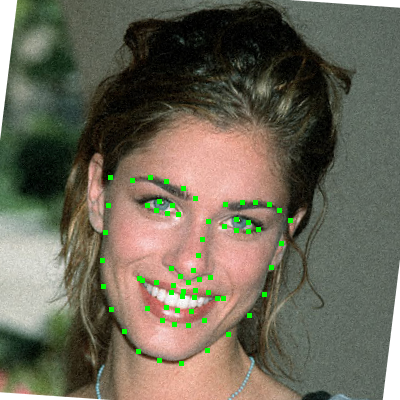}&
\includegraphics[width=0.15\linewidth]{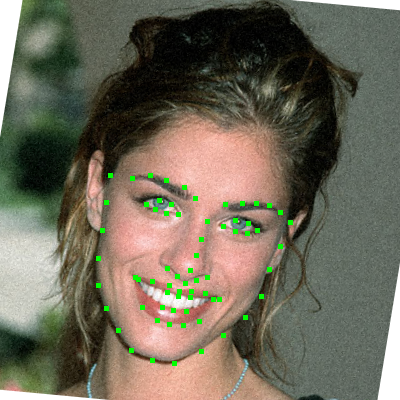}
\end{array}$
\end{center}
\caption{Data augmentation. This sequence of images continuously changes in brightness, Gaussian noise, scale, and projective distortion. In this way we can augment a single image into a ``pseudo" video.}
\label{pics:2daug}
\end{figure}

\subsection{Results on Public Datasets}

The comparison of our result with previous methods on the three test sets of 300-VW is shown in Table \ref{tab:300vw}. Both the base model described in Section \textit{Experimental Setup} and our new model, which is retrained using L2 loss with our proposed method, are compared in this table. The better performance of our new model shows that the noise reduction can improve the performance on all three subsets. Compared with the baseline model, the noise reduction training provides our network with a better comprehension of the data distribution in hyperspace.

\begin{table*}[htbp]
\caption{Table 2. Comparison of NME ($\%$) on 300-VW dataset}
  \label{tab:300vw}
  \begin{center}
  \begin{tabular}{l c c c c }
    \hline
    Method & Training Data & Subset-1 & Subset-2 & Subset-3\\
    \hline
    \multicolumn{5}{c}{Inter-pupil distance}  \\ \hline
    SDM \cite{xiong2013supervised}& / & 7.41 &	6.18 &	13.04 \\
    TCDCN \cite{zhang2016learning} & 300-W & 7.66 &	6.77 &	14.98 \\
    CFSS \cite{zhu2015face} & 300-W & 7.68 &	6.42 &	13.67 \\
    DRSN \cite{miao2018direct} & AFLW, 300-W, CelebA, MAFL, 300-VW & 5.33 & 4.92 & 8.85 \\
    \hline
    \multicolumn{5}{c}{Inter-ocular distance}  \\ \hline
    \textbf{Ours-base} & 300-W & 5.13 &	5.94 &	8.81 \\
    \textbf{Ours-new} & 300-W & \textbf{4.60} &	\textbf{4.04} &	\textbf{8.49}\\
    
  \hline
\end{tabular}
 
\end{center}
\end{table*}

To keep consistency with previous works, we also report the NME on the three test sets of 300-W in Table \ref{tab:300w}. When applied to a single image, our method only uses the module labeled ``detection" in Figure \ref{fig:framework}.
The best result on 300-W so far is reported by LAB \cite{wayne2018lab} trained with a new dataset WFLW including 10,000 images with different environments, poses, occlusions, etc. In addition, SBR \cite{dong2018supervision} and SAN \cite{dong2018style} also include private training sets. Our model is able to reach comparable results with only 300-W's training set.

\begin{table*}[htbp]
\caption{Table 3. Comparison of NME ($\%$) on 300-W dataset}
  \label{tab:300w}
\begin{center}
\begin{tabular}{l c c c c }
    \hline
    Method & Training Data & Common & Challenging & Full Set\\
    \hline
    \multicolumn{5}{c}{Inter-pupil distance}  \\ \hline
    SDM \cite{xiong2013supervised} & / & 5.57 & 15.40 & 7.52 \\
    LBF \cite{ren2014face} & 300-W & 4.95 & 11.98 & 6.32 \\
    MDM \cite{trigeorgis2016mnemonic} & 300-W & 4.83 & 10.14 & 5.88 \\
    TCDCN \cite{zhang2016learning} & MAFL & 4.90 & 8.60 & 5.54 \\
    CFSS \cite{zhu2015face} & 300-W & 4.73 & 9.98 & 5.76 \\
    DRA-STR \cite{lv2017deep} & 300-W, AFLW & 4.36 &	7.56 &	4.99 \\
    DRSN \cite{miao2018direct} & AFLW, 300-W, CelebA, MAFL, 300-VW & 4.12 & 9.68 & 5.21 \\
    3DALBF \cite{guo2019face} & 300-W, 300W-LP & 3.69 & 10.03 & 4.93 \\
    LAB \cite{wayne2018lab} & WFLW & 3.42 & 6.98 & 4.12 \\
    \hline
    \multicolumn{5}{c}{Inter-ocular distance}  \\ \hline
    SBR \cite{dong2018supervision} & 300-W, AFLW, 300-VW & 3.28 & 7.58 & 4.10 \\
    SAN \cite{dong2018style} & 300-W, AFLW & 3.34 & 6.60 & 3.98 \\
    LAB \cite{wayne2018lab} & WFLW & \textbf{2.98} & \textbf{5.19} & \textbf{3.49} \\
    \textbf{Ours} & 300-W &3.62 & 5.41 & 3.97 \\
  \hline
\end{tabular}

\end{center}
\end{table*}


\begin{figure}
  \includegraphics[width=\linewidth]{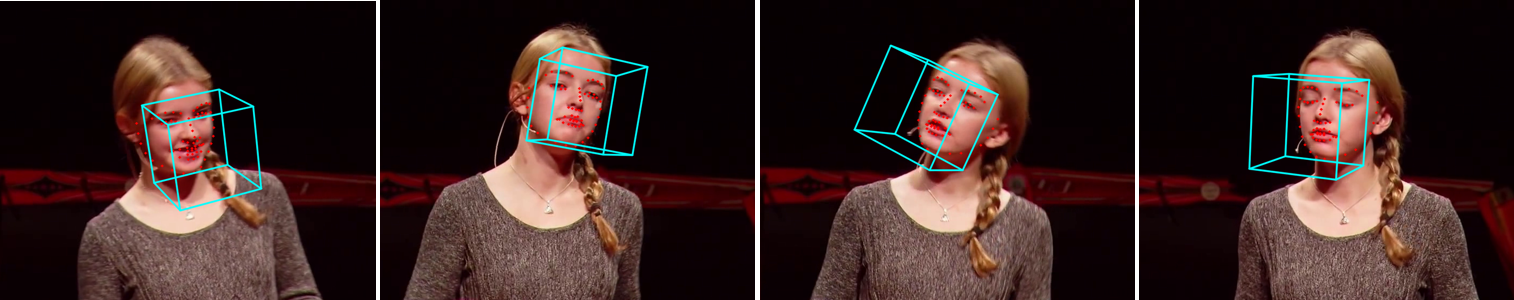}
  \caption{Video frames from the 300-VW dataset with the detected 3D bounding box and 68 landmark points}
  \label{fig:teaser}
\end{figure}

\subsection{Effect of Noise on the Accuracy and the Precision}

\begin{figure}[htbp]
\begin{center}
\includegraphics[width=\linewidth]{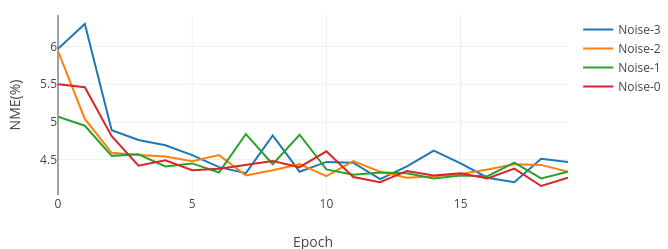}
\end{center}
   \caption{NME ($\%$) of models trained with different amounts of injected noise. \textmd{The $X$-axis is the epoch of training. Each line represents a series of models acquired by injecting a fixed amount of Gaussian noise to facial landmark locations in the training set. All models are tested on the same 300-W test set. As the epoch increases, all models converge to a similar level of accuracy regardless of how much noise has been injected in the training data.}}
\label{fig:addnoise}
\end{figure}

As discussed in Section \textit{Noise Modelling}, we can estimate the mean of the target landmark values to any desired degree of accuracy if given a sufficiently large and representative training set \cite{white1990connectionist}. L2 loss can guarantee a high accuracy towards the ``true and only" target as long as the noise is zero-mean. In order to verify this, we add different amounts of Gaussian noise (from 0\% to 3\%) following the settings in \cite{dong2018supervision} to the training set and train another three models using the same approach as for the base model. The results in Figure \ref{fig:addnoise} show that even if these models have never seen clean data, they are able to reach the same level of accuracy as the base model. Similar results are also discussed in previous papers \cite{dong2018supervision}\cite{lehtinen2018noise2noise}.  

When we tried to fine-tune our base model on a small subset of challenging images, we also discovered a possible ``blessing" of the noise. Since the challenging images are of a small number and severely deviate from the main set, the fine-tuning can lead to a generally worse result on the 300-VW test set because of over-fitting on the challenging set. We compare the fine-tuning results from original data, augmented data (higher probability of over-fitting), and data with 3\% additive Gaussian noise in Table \ref{tab:300vw-noise}. Surprisingly, the model trained with additive noise performs better than the other two. This result may imply the positive effect of known noise in overcoming the trend of over-fitting. 

\begin{table}[htbp]
  \caption{Table 4. Comparison of NME ($\%$) on 300-VW test set}
  \label{tab:300vw-noise}
  \begin{center}
  \begin{tabular}{l c c c}
    \hline
    Training data & Subset-1 & Subset-2 & Subset-3\\
    \hline
    Original & 5.62 &	4.63 &	9.97\\
    Augumented & 5.88 &	4.76 &	9.98 \\
    Noisy & 5.55 &	4.54 &	9.87 \\
  \hline
\end{tabular}
\end{center}

\end{table}

\begin{figure}[htbp]
    \includegraphics[width=\linewidth]{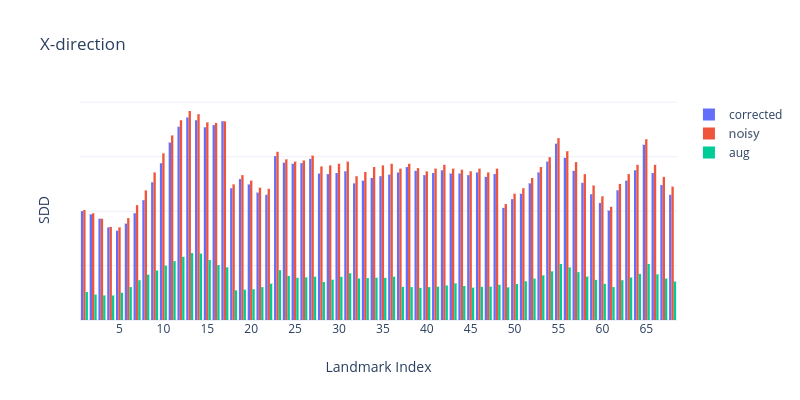}
    \includegraphics[width=\linewidth]{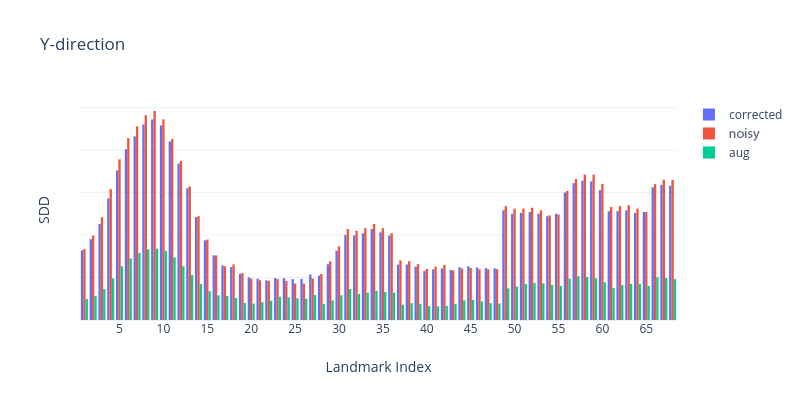}
    \caption{Comparison of SDD of different models. \textmd{In this graph, the horizontal axis denotes the landmark point's index(from 1 to 68), and the vertical axis is the standard deviation of difference (SDD) as defined in Section \textit{Noise Modelling}. The horizontal direction and vertical direction results are plotted in the top and bottom graphs, where different color bars are results of three different models: corrected (trained on corrected data), noisy (trained on corrected data with injected noise), and aug (trained on augmented corrected data).}}
    \label{fig:stab}
\end{figure}

The annotation noise's curse to the detection accuracy has already been discussed in earlier sections. Besides, our experimental results also reveal that the noise in the training set can degrade model's precision and lead to more jitters in the video results. We compare the facial landmark outputs of three models trained with different data: corrected data, noisy data, and augmented data. As shown in Figure \ref{fig:stab}, compared with correted-data model, the noisy-data model is higher in SDD. It indicates that the injected noise will increase the spatial instability of output landmarks. Besides, the augmented-data model shows the best performance among all three models by a large margin, which indicates our 2d-augmentation method can reduce the output jitters without extra modifications to the model.  

\section{Conclusion}
In this paper, we investigate the effect of noise in facial landmark detection task by modeling the noise in both the training set and detection output, and comparing results of models trained with different noise and training strategies. Our results show the great potential of getting better landmark detectors trained on public datasets with our proposed noise reduction method. Our method is capable of handling multiple types of noise in both the annotation and detection processes. Besides, we also discuss the relationship between the loss function and different types of noise. Our further experiments suggest that the noise injection could be a good method to avoid over-fitting.

\small
\bibliography{egbib}
\bibliographystyle{plain}






\end{document}